\documentclass[letterpaper, 10 pt, conference]{ieeeconf}  

\usepackage{amsmath}
\IEEEoverridecommandlockouts                              

\usepackage{graphicx}
\usepackage{algorithm}
\usepackage{algorithmic}
\usepackage{mathtools}
\usepackage{xcolor}
\usepackage{subfig}

\title{\LARGE \bf
Trajectory Optimization for Robust Humanoid Locomotion with Sample-Efficient Learning*}

\author{Majid Khadiv$^{1}$, Mohammad Hasan Yeganegi$^{2}$, S. Ali A. Moosavian$^{2}$, Jia-Jie Zhu$^{1}$ and Ludovic Righetti$^{1,3}$
\thanks{*This work is supported by New York University, the Max-Planck Society, the European Union’s Horizon 2020 research and innovation program (grant agreement No 780684 and European Research Council’s grant No 637935) and the National Science Foundation (grant CMMI-1825993)}
\thanks{$^{1}$ Max Planck Institute for Intelligent Systems, Tuebingen, Germany. {\tt\small firstname.lastname@tuebingen.mpg.de}}%
\thanks{$^{2}$ K. N. Toosi University of Technology, Tehran, Iran
        {\tt\small yeganegi.m.h@email.kntu.ac.ir}
        {\tt\small moosavian@kntu.ac.ir}}%
\thanks{$^{3}$ Tandon School of Engineering, New York University, New York, USA. {\tt\small ludovic.righetti@nyu.edu}}%
}

\begin{document}
\maketitle
\thispagestyle{empty}
\pagestyle{empty}

\begin{abstract}

Trajectory optimization (TO) is one of the most powerful tools for generating feasible motions for humanoid robots. However, including uncertainties and stochasticity in the TO problem to generate robust motions can easily lead to an interactable problem. Furthermore, since the models used in the TO have always some level of abstraction, it is hard to find a realistic set of uncertainty in the space of abstract model. In this paper we aim at leveraging a sample-efficient learning technique (Bayesian optimization) to robustify trajectory optimization for humanoid locomotion. The main idea is to use Bayesian optimization to find the optimal set of cost weights which compromises performance with respect to robustness with a few realistic simulation/experiment. The results show that the proposed approach is able to generate robust motions for different set of disturbances and uncertainties.

\end{abstract}

\section{INTRODUCTION}
Since humanoid robots are inherently both redundant (so many degrees of freedom in the limbs' structure) and underactuated (floating base without direct actuation), generating feasible and optimal motions for them is very challenging. Trajectory optimization (TO) is a strong tool to take into account all the physical and geometrical constraints and at the same time yield the optimal motion minimizing the cost function. However, the discrepancy between the model used in the TO problem and the real robot, as well as uncertainties in the constraints set make the generated motion fragile. 

One principled way to deal with this problem is to use stochastic or robust TO approaches to add the notion of uncertainties to the problem. Although systematic, this approach suffers from two problems: 1) identifying different types of uncertainties and projecting them to a realistic set of constraints is challenging and 2) adding stochastic uncertainties can easily lead to an intractable problem, and in most cases can be solved only for simplified worst case scenarios. 

\begin{figure}[!t]
    \centering
    \includegraphics[clip,trim=3cm 4cm 7cm 4cm,width=8cm]{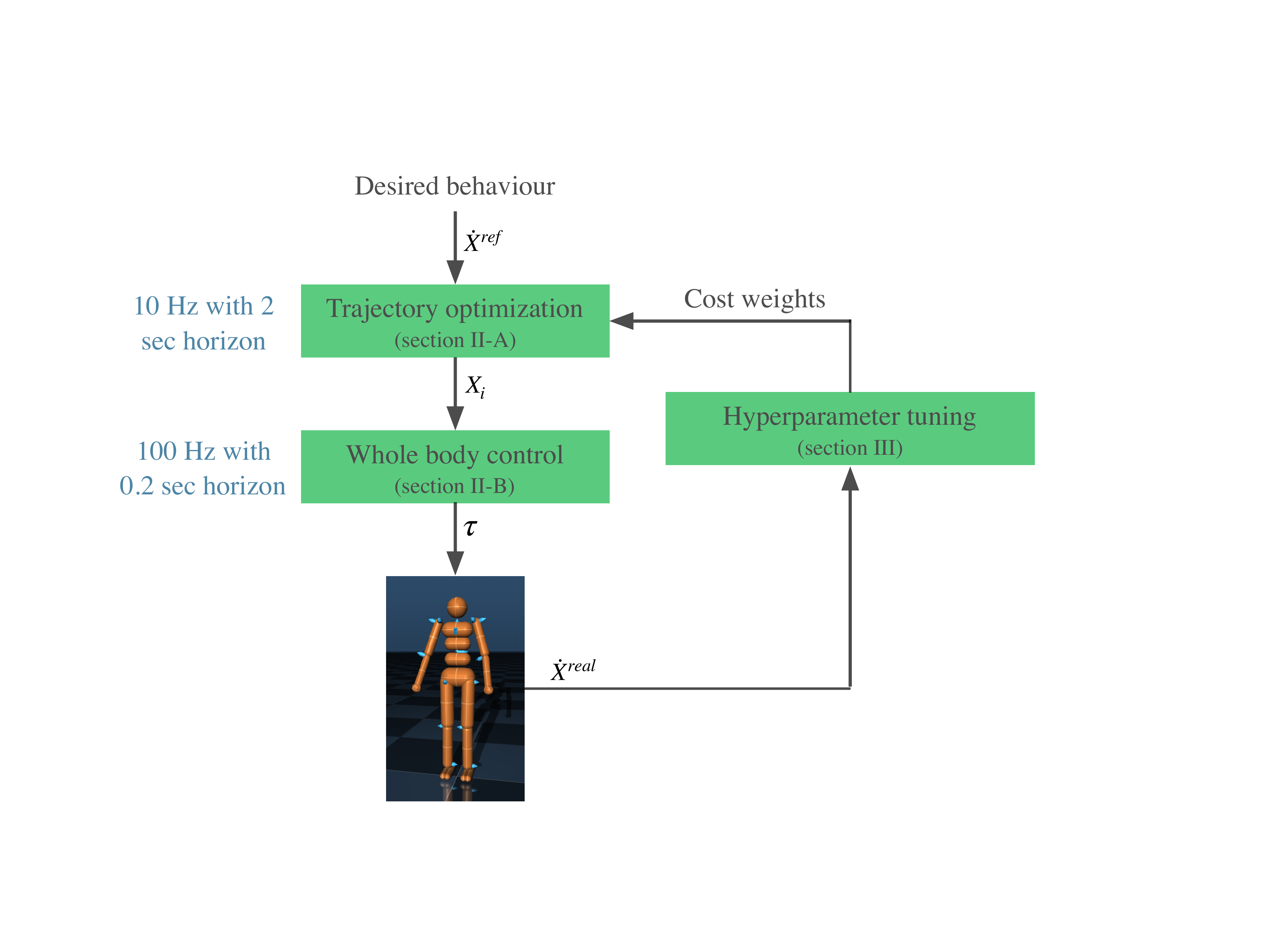}
    \caption{A high-level block diagram of the proposed approach}
    \label{fig:block_diagram}
    \vspace{-5 mm}
\end{figure}

\begin{figure}[!t]
    \centering
    \subfloat{\includegraphics[width=1.4cm]{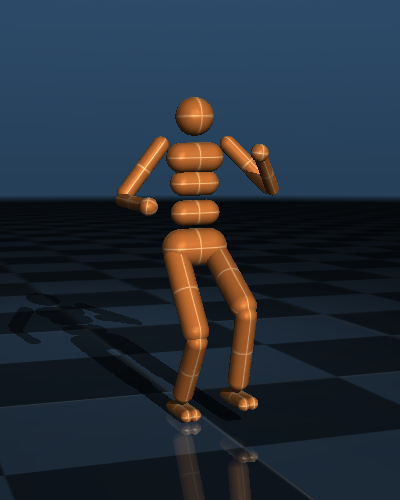}}
    \subfloat{\includegraphics[width=1.4cm]{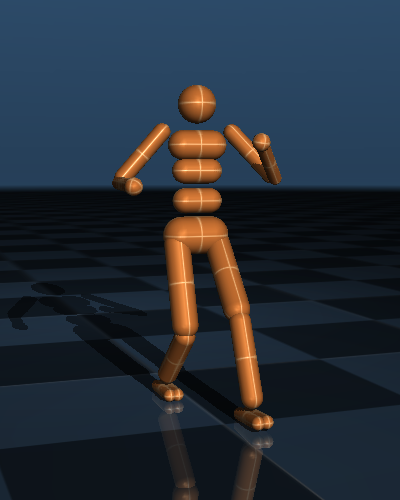}}
    \subfloat{\includegraphics[width=1.4cm]{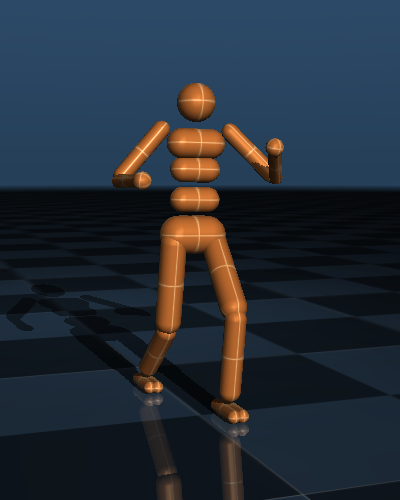}}
    \subfloat{\includegraphics[width=1.4cm]{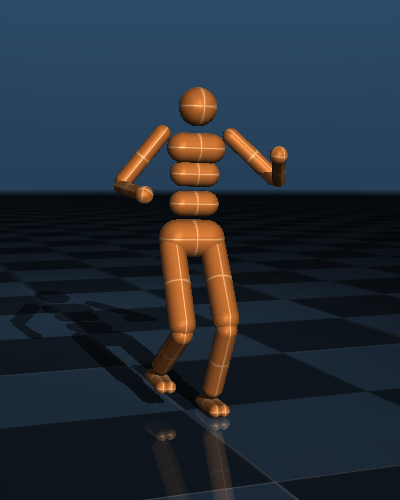}}
    \subfloat{\includegraphics[width=1.4cm]{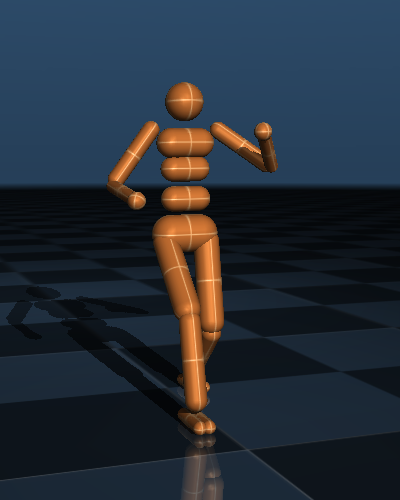}}
    \subfloat{\includegraphics[width=1.4cm]{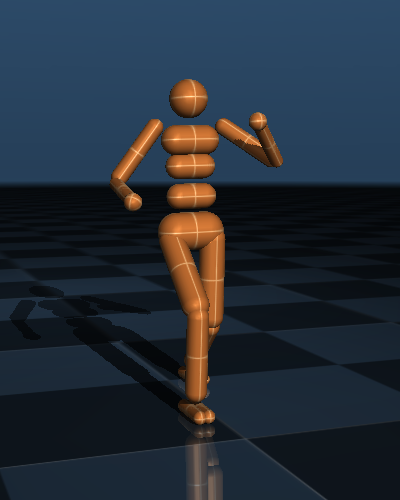}}\\
    \subfloat{\includegraphics[width=1.4cm]{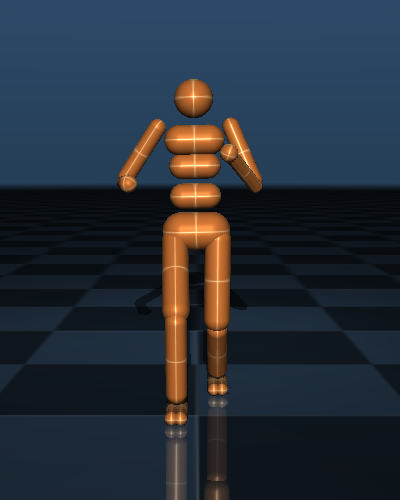}}
    \subfloat{\includegraphics[width=1.4cm]{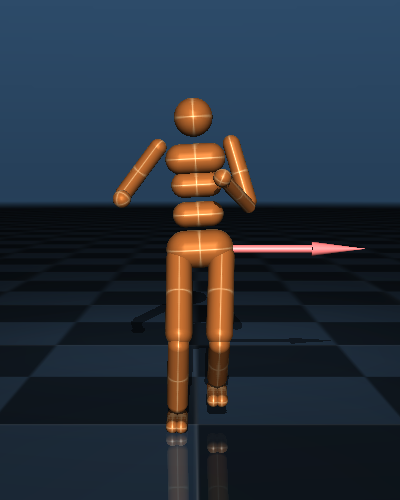}}
    \subfloat{\includegraphics[width=1.4cm]{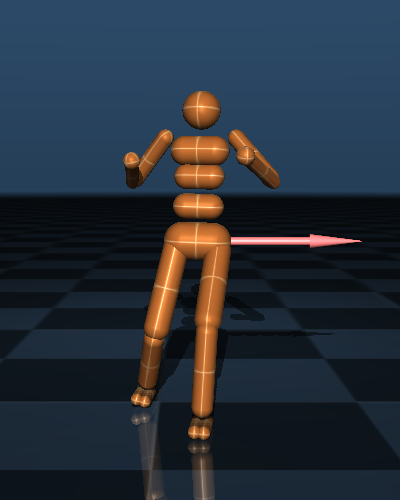}}
    \subfloat{\includegraphics[width=1.4cm]{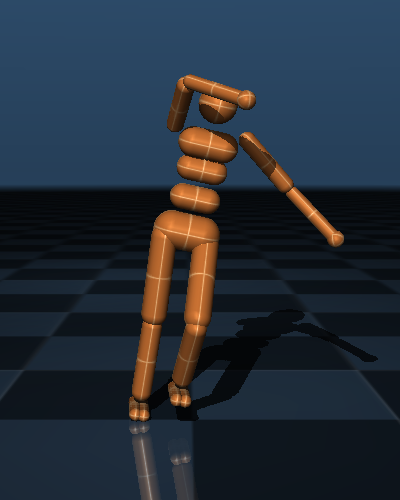}}
    \subfloat{\includegraphics[width=1.4cm]{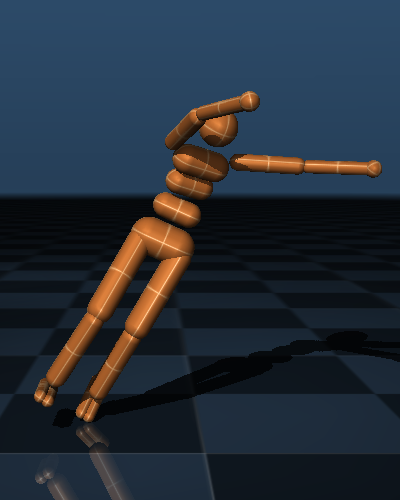}}
    \subfloat{\includegraphics[width=1.4cm]{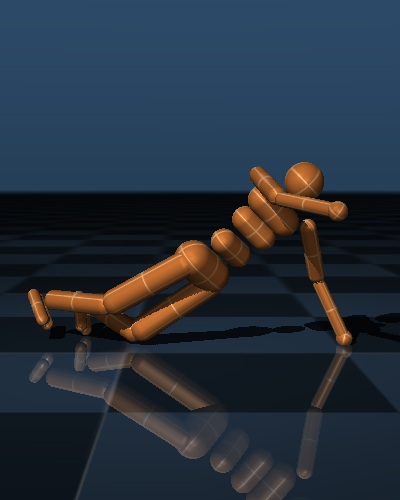}}
    \caption{Screenshots of simulation in MuJoCo, (top) without disturbance (bottom) with lateral push}
    \label{screenshot}
    \vspace{-5 mm}
\end{figure}

The main contribution of this work is to combine the strength of trajectory optimization and sample-efficient learning (Bayesian optimization) to generate robust motion for different kinds of uncertainties with a low number of experiments. Contrary to the available approaches for generating bipedal locomotion using BO \cite{rai2018bayesian,calandra2017bayesian}, we propose to use constrained gradient-based approaches to generate feasible motion using the robot abstract model, and then use BO to tune the cost parameters of the TO problem in the presence of disturbances during simulation of the full robot. By proper formulation of the problem in this setting, we can trade-off robustness against performance, given a set of realistic disturbances/uncertainties in the full robot simulation. Fig. \ref{fig:block_diagram} shows the block diagram of our approach.  

\section{Optimal control problem}\label{section:OC}
\subsection{First Stage : Convex Trajectory Optimization for Walking}
We use the TO approach proposed in \cite{khadiv2017pattern} which is an extended version of a standard walking pattern generator \cite{herdt2010online} to compute center of mass (CoM) motion and footstep locations to achieve a desired walking velocity. In this approach, walking is formulated as a trade-off among three main cost terms: one for the task goal (desired velocity tracking) and two to increase motion robustness (foot tip-over avoidance and slippage avoidance). 
\begin{align}
    \label{eq:TO}
    \underset{\dddot X_i , X_i^f}{\text{min.}} \, \sum\limits_{i=1}^{N}\, & \alpha \Vert \dot X_{i}-\dot X_{i}^{ref} \Vert^2  + \beta \Vert Z_{i}-Z_{i}^{ref} \Vert^2  + \gamma \Vert \mu_{i} \Vert^2  \nonumber\\
    \text{s.t.} \qquad &\mu_{i} \in friction\,cone \quad , \quad \forall i=1,...,N. \nonumber\\
    \qquad &X_i^f \in reachable\,area \quad , \quad \forall i=1,...,N. \nonumber\\
     \qquad &Z_{i} \in support\,polygon \quad , \quad \forall i=1,...,N.
\end{align}
where $X=[c_x,c_y]^T$ is the CoM position in horizontal plane. $Z=[z_x,z_y]^T$ is the zero moment point (ZMP) position and $\mu$ is the required coefficient of friction (RCoF). $\dot{X}^{ref}$ is the desired walking velocity, $Z^{ref}$ is the desired ZMP which is considered at the center of the foot to generate maximum feasibility margin. As it is shown in \cite{khadiv2017pattern}, this optimization problem can be written as a quadratic program, assuming linear inverted pendulum dynamics and a polyhedral approximation of the friction cone. 

\subsection{Second Stage : iLQG for generating whole body torques}
We use an iLQG controller to map the desired CoM and feet trajectories to whole body torques, taking into account box inequality constraints on controls \cite{tassa2014control}. We use iLQG (with a short horizon of 0.2 sec) as a whole body controller to track the desired trajectories from the first stage. For our walking problem, we use a 27-DoF humanoid robot model and the iLQG computes joint torques every 0.01 sec.

\section{Computing cost weights for robust humanoid locomotion} \label{section:BO}
We propose to close the loop of the system and automatically optimize the cost function of the abstract pattern generator to find plans that are robust to full robot dynamics and environment uncertainties. 
We formulate an overall optimization problem based on the quantities in Fig. \ref{fig:block_diagram} (and solve it using BO):
\begin{align}
\label{eq:opt_final}
\underset{\delta}{\text{min.}} & \, J(\delta)\vcentcolon = \sum\limits_{i=1}^{N}\, \Vert \dot X^{\text{real}}_{i}(\delta) -\dot X_{i}^{des} \Vert^2 + \lambda \phi(h_N^\delta),\\
\text{s.t.} \qquad & \dot X^{\text{real}}_{i}(\delta)\, \text{is the output of the plant (robot) in Fig~\ref{fig:block_diagram}. \nonumber}
\end{align}
$\delta=(\alpha, \beta, \gamma)$ is the collection of the hyper-parameters used in optimization problem \eqref{eq:TO}. We set $\alpha=1$ to guarantee the viability of the gait \cite{herdt2010online}, and optimize for $\beta$ and $\gamma$ to find the best trade-off between robustness and performance, where robustness depends on the environment. $\dot X^{\text{real}}_{i}(\delta)$ is the measured CoM velocity resulting from applying the control (i.e. solving the QP in \eqref{eq:TO} and applying iLQG tracking) in simulation under unknown disturbances. 
$h^\delta_N$ is CoM height at terminal time. $\lambda$ is a trade-off parameter which is straight forward to set. $\phi$ penalizes robot falling, e.g., $\phi= \max (|h^\delta_N-h_{des}|-\text{threshold}, 0)$.

\section{Preliminary results}\label{section:results}
In this section we show that our approach can adapt cost functions to generate robust walking gaits in the presence of various disturbances. The range of weight values we consider is $0 \leq \beta,\gamma \leq 1000$. In all cases we start with $\beta,\gamma =1000$ where we have high ZMP and RCoF margins. We investigate four cases, i. e. (a) without disturbance, (b) with external pushes on upper body, (c) unknown decrease of the surface friction coefficient, and (d) both external forces and decrease of friction coefficient.

\begin{figure}[!t]
    \centering
    \subfloat[Original cost]{\includegraphics[width=4.5cm]{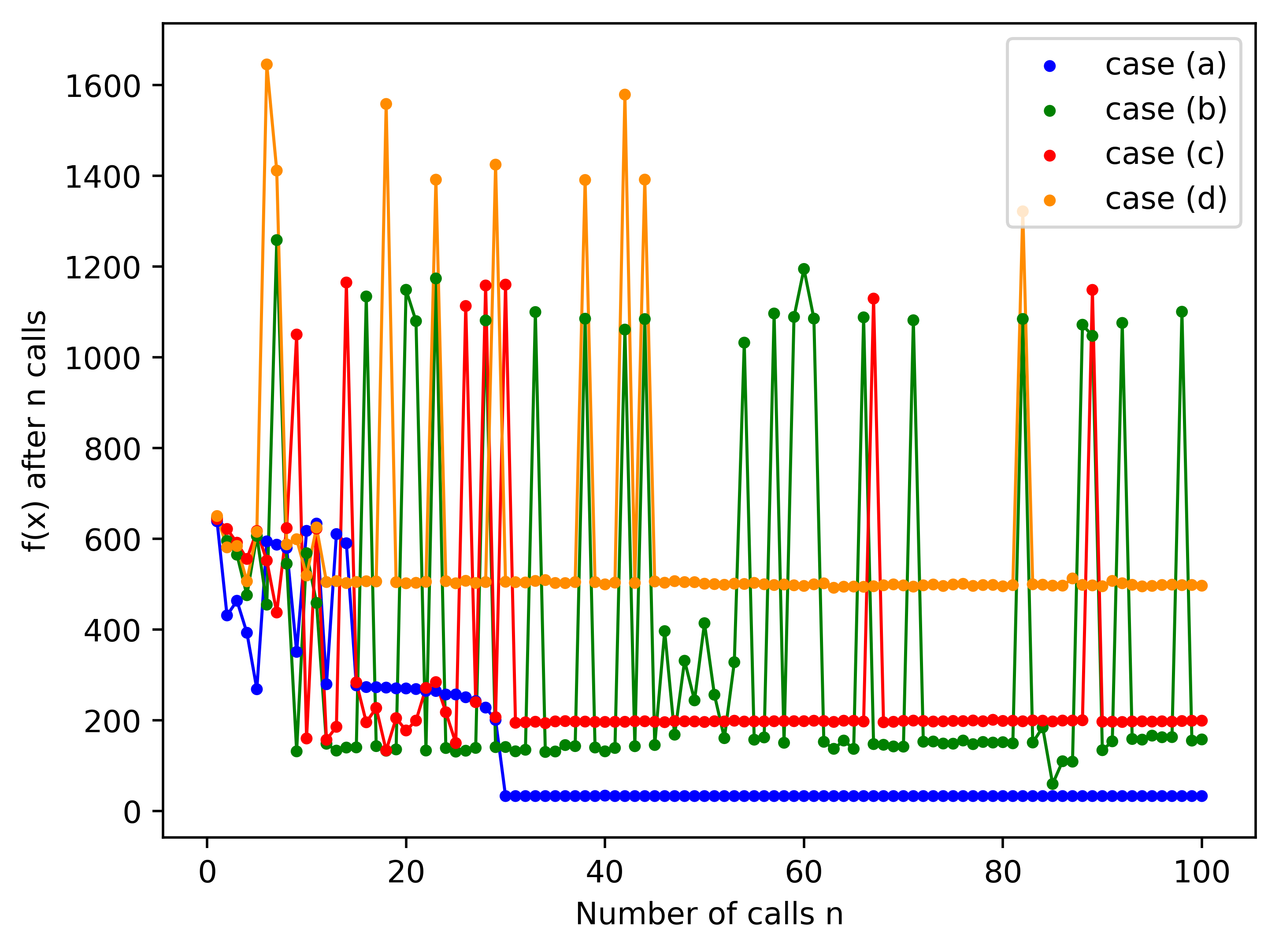}\label{original_cost}}
    \subfloat[minimum cost after n call]{\includegraphics[width=4.5cm]{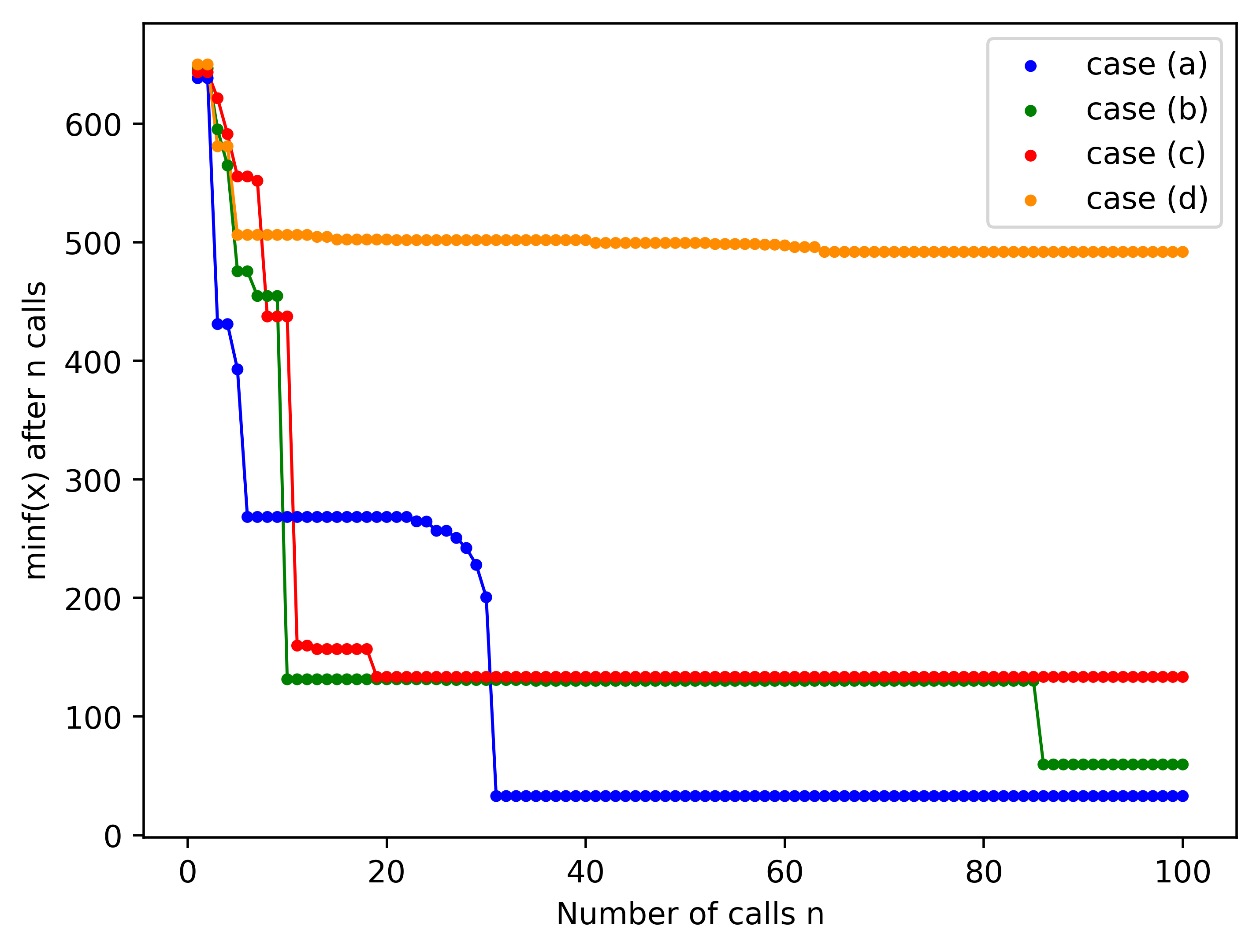}\label{minimum_cost}}
    \caption{History (path) of the exploration by BO}
    \label{cost}
    \vspace{-7 mm}
\end{figure}

In Fig. \ref{cost}\subref{original_cost} we plot the evolution of cost values of BO at each iteration for all cases of this scenario. Large variations of the values correspond to the high penalty given to the failed cases. As expected, the decrease in the cost is not monotonic, as we are not using gradient-based optimization. In Fig. \ref{cost}\subref{minimum_cost}, we plot the minimum value of the current cost and all last calls for evaluating the function. The optimal values for each case are: (a) $\beta=0, \gamma=0$, (b) $\beta=400.11, \gamma=1.71$, (c) $\beta=27.09, \gamma=51.64$ (d) $\beta=778.5, \gamma=188$. Interestingly, for all the cases after a few iterations (around 15 calls for cases (b), (c), (d), and after 25 calls for case (a)), the cost has already settled. This suggests that within a few experiments the cost function can be optimized and lead to robust walking in uncertain environments, which is important for deployment on real humanoid robots where experiments can be time consuming.

\section{Ongoing and future work}\label{section:conclusion}
These results are a proof of concept where we tuned two cost variables using BO. We are currently applying the concept to more complicated problems: using this approach to optimize more complex costs weights for RCoF and ZMP in sagittal and lateral directions as well as costs on step locations being far from boundaries of reachable area. 
A longer term goal of this project is to apply the approach to more complicated TO problems based on centroidal dynamics and full-body dynamics.

\bibliography{Master}
\bibliographystyle{IEEEtran}

\end{document}